\documentclass[10pt,twocolumn,letterpaper]{article}

\usepackage{wacv}
\usepackage{times}
\usepackage{epsfig}
\usepackage{graphicx}
\usepackage{amsmath}
\usepackage{amssymb}
\usepackage{setspace}
\usepackage{multirow}
\usepackage{float}
\usepackage{caption}
\usepackage{authblk}
\def\wacvPaperID{62} 

\wacvfinalcopy 

\ifwacvfinal
\def\assignedStartPage{1} 
\fi

\ifwacvfinal
\usepackage[breaklinks=true,bookmarks=false]{hyperref}
\else
\usepackage[pagebackref=true,breaklinks=true,colorlinks,bookmarks=false]{hyperref}
\fi

\ifwacvfinal
\else
\fi

\begin{document}

\title{SIDE: Center-based Stereo 3D Detector with
Structure-aware \\ Instance Depth Estimation}


\author[1]{Xidong Peng}
\author[2]{Xinge Zhu}
\author[2]{Tai Wang}
\author[1]{Yuexin Ma}
\affil[1]{ShanghaiTech University}
\affil[2]{The Chinese University of Hong Kong \authorcr {\tt\small \{linmo1533, zhuxinge123, taiwang.me\}@gmail.com, mayuexin@shanghaitech.edu.cn}}

\maketitle
\definecolor{Gray}{gray}{0.5}
\newcommand{\demph}[1]{\textcolor{Gray}{#1}}

\def\algorithmname{SIDE}

\begin{abstract}

3D detection plays an indispensable role in environment perception. Due to the high cost of commonly used LiDAR sensor, stereo vision based 3D detection, as an economical yet effective setting, attracts more attention recently. For these approaches based on 2D images, accurate depth information is the key to achieve 3D detection, and most existing methods resort to a preliminary stage for depth estimation. They mainly focus on the global depth and neglect the property of depth information in this specific task, namely, sparsity and locality, where exactly accurate depth is only needed for these 3D bounding boxes. Motivated by this finding, we propose a stereo-image based anchor-free 3D detection method, called structure-aware stereo 3D detector (termed as \algorithmname), where we explore the instance-level depth information via constructing the cost volume from RoIs of each object. Due to the information sparsity of local cost volume, we further introduce match reweighting and structure-aware attention, to make the depth information more concentrated. Experiments conducted on the KITTI dataset show that our method achieves the state-of-the-art performance compared to existing methods without depth map supervision.
\end{abstract}

\section{Introduction}

\begin{figure}[]
	\centering
	\includegraphics[scale=0.225]{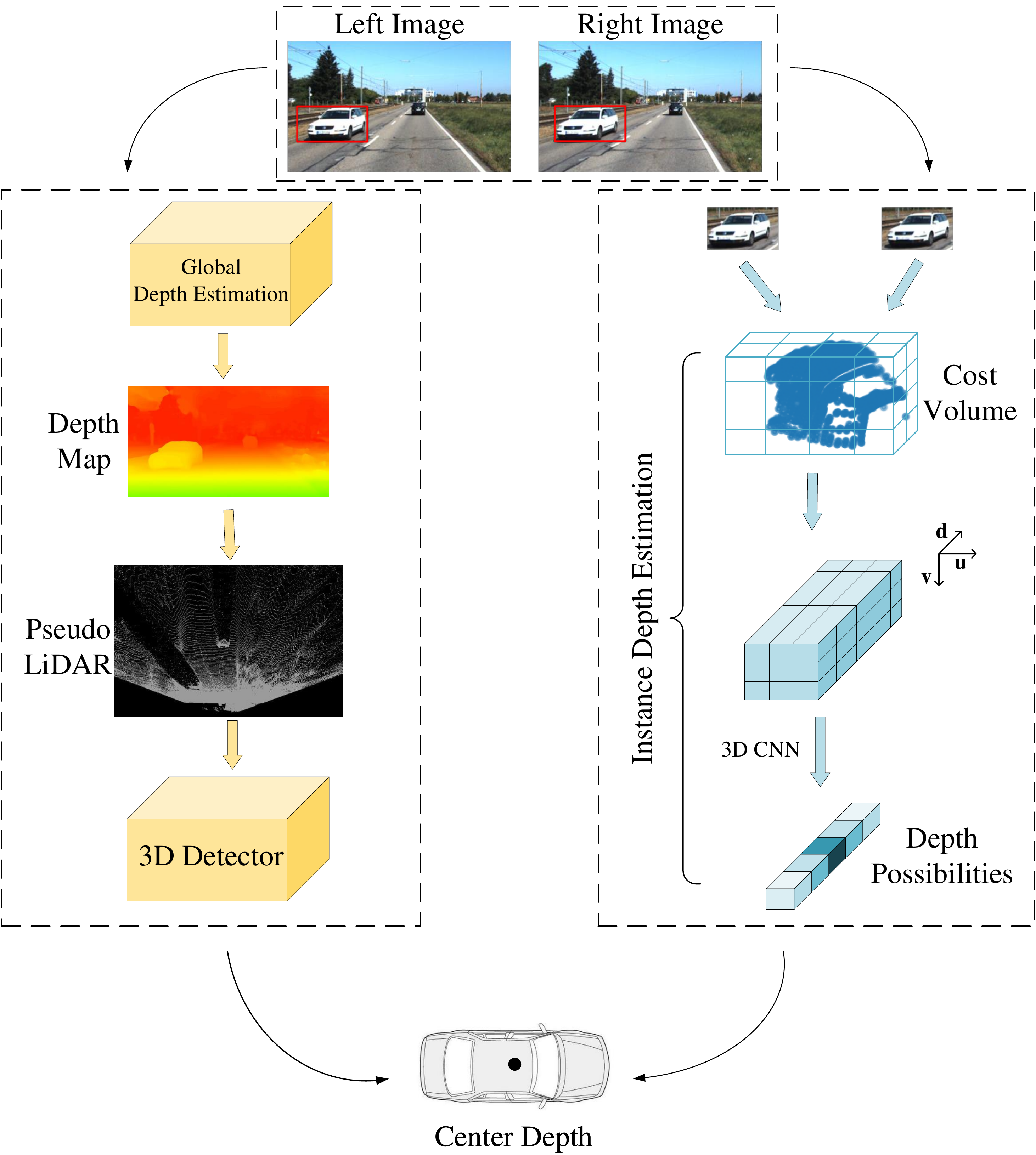}
	\caption{The left shows global depth estimation used in previous work while the right is instance depth estimation used in our work. Our depth estimation is only for objects' center and does not need depth map for supervision.}
	\label{fig_teaser}
\end{figure}

3D object detection is important for scene understanding and widely applied in many applications, such as autonomous driving \cite{no19, 2018CubeSLAM, ma2019trafficpredict} and virtual reality. The rapid progress of 3D detectors have been witnessed in recent years and most state-of-the-art approaches leverage the data collected by LiDAR \cite{li2016vehicle, no3, no5, no18, no13} considering it can provide accurate 3D information. However, LiDAR is expensive to deploy or maintain in practical use, and has limited sensing range in some cases, which makes the vision-based 3D detection methods draw more attention recently. Compared to LiDAR-based methods, the key information about depth is not directly manifest in the given 2D images. This problem is especially prominent for monocular cases \cite{no19, no12, no20, no11, no22}. In comparison, binocular methods could leverage the stereo geometry to physically compute the depth, and thus are more reasonable approaches to detecting 3D objects from 2D images. Therefore, we target at the key problem of depth estimation and aim at proposing an efficient and effective stereo-based 3D detector.


Previous stereo-based work~\cite{pl, no37, no25} typically estimates the global depth field to assist subsequent 3D detection. Although it achieves promising performance even close to LiDAR-based methods, there are several limitations. First, it is time-consuming due to the computational overheads introduced by per-pixel dense estimation. In addition, considering most of the area in the image belongs to the background, a better performance of global estimation sometimes needs a trade-off from foreground regions, which are actually more important in this specific task. Furthermore, it needs the dense depth labels for supervision, and thus brings extra annotation costs. Hence, instead of estimating the depth globally, focusing on the depth estimation of instances with only 3D bounding boxes labels can be more effective in stereo-based 3D detection.


In this paper, we propose a novel 3D detection method, \ie, \algorithmname, to solve above problems, in which it performs depth estimation only for objects' center with corresponding 2D regions of interest (RoI). Compared with previous methods, the dense depth labels are not required and running time can be reduced with the simplification of required depth estimation. Fig.\ref{fig_teaser} shows the comparison between our work and previous work intuitively. During depth estimation, we first introduce match-reweight strategy to take advantage of the internal similarity of cost volume. Since instance depth information is sparse in space and the position of object's 3D center is not directly represented in the image, we introduce structure-aware attention mechanism to extract the structural information of local patch in the front view and the bird’s eye view by convolution, then condense the information into the original cost volume to make the depth feature more concentrated. Based on the accurate depth estimation, we further propose a simple and efficient post-processing under the geometric constraint to refine our detection results. 

We evaluate the proposed SIDE method with KITTI 3D dataset~\cite{no1}. Specifically, our $AP_{3D}$ of car category is better than the state-of-the-art methods IDA-3D~\cite{ida3d} in all kinds of cases with IoU=0.7. Especially, in the moderate and hard case, our method performs better than IDA-3D with over 3\% $AP_{3D}$, which means our method can better detect objects far away or with large occlusions.

The contributions of our proposed method are mainly summarized as follows.

\begin{itemize}
  \item [1)] 
We investigate the natural property of depth information in stereo 3D detection, namely, sparsity and locality, and reroute the global depth to structure-aware instance depth.
  \item [2)] 
We introduce a novel stereo 3D detector by accurately predicting the center depth of each object with the match-reweight and structure-aware attention. 
  \item [3)]
A simple yet effective post-processing method is proposed to refine the detection results under the geometric constraint.
  \item [4)]
Evaluated on the KITTI 3D dataset, we achieve state-of-the-art performance compared with the stereo-based methods without depth map supervision.
\end{itemize}


\begin{figure*}
	\centering
	\includegraphics[scale=0.3]{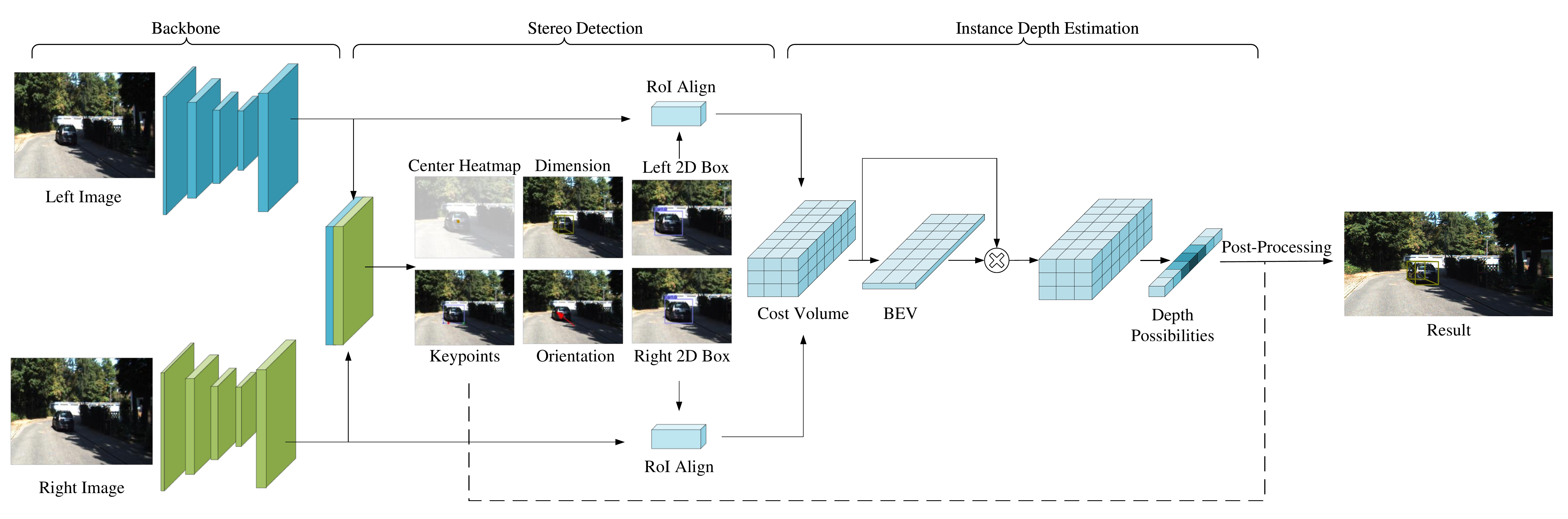}
	\caption{Network architecture of the proposed \algorithmname. It takes only the stereo RGB images as inputs. After the backbone network, the network outputs 7 branches which predict center depth, 2D center heatmap, stereo 2D boxes, object’s kepoints, dimension and steering angle of 3D box based on the idea of anchor-free. Center depth branch is different from others, in which we perform instance depth estimation by selecting 2D RoI of objects. In order to get accurate depth, we introduce match-reweight and structure-aware attention in the depth estimation to make depth feature more focused. Finally, detection results are further refined through the post-processing.}
	\label{fig1}
\end{figure*}

\section{Related Work}


\noindent\textbf{LiDAR based methods}\quad Most of the state-of-the-art 3D detection methods rely on LiDAR, because it provides accurate depth information of objects. These methods process LiDAR data in different representations. \cite{vote3deep, no15, no31, zhu2020ssn, wang2020reconfig, zhu2021cylindrical1,zhu2021cylindrical2} utilize structured voxel representation to quantize the LiDAR data and feed them into 2D or 3D CNN to detect 3D object, while \cite{no5, no18, no13} project the LiDAR data into 2D bird’s eye view or front view representations. Instead of transforming the representations of point cloud, \cite{no23, no3} directly takes raw point cloud as input to localize 3D object based on the frustum region. Additionally, the idea of anchor-free is applied to the LiDAR method in \cite{no8, no30}, which reduces the time consuming for 3D detection. Although the performance of LIDAR-based methods is superior, compared with the high cost of LIDAR, the image-based methods are more practical currently. 

\noindent\textbf{Moncular image based methods}\quad Because monocular cameras are cheaper than LiDAR or stereo cameras, monocular-based 3D object detection naturally becomes a hot spot for both industry and academia. \cite{no19, no12, no20, wang2021fcos3d, wang2021pgd} extend the state-of-the-art 2D object detector to regress the orientation and dimensions of the object’s 3D bounding box. \cite{no11, no22} explicitly utilize sparse information by predicting series of keypoints of regular-shape vehicles, then the 3D object pose can be constrained by wireframe template fitting. \cite{no17} predicts the nine perspective key points of the 3D bounding box, and uses geometric constraint to recover the three-dimensional information of the object. \cite{no6} considers the relationship between paired samples to improve monocular 3D target detection. These methods are cheap and fast, but the performance is not satisfying because it is difficult to obtain accurate depth information, which is very critical for 3D detection.

\noindent\textbf{Stereo image based methods}\quad Stereo cameras are much cheaper than LiDAR and stereo images can provide depth information implicitly through the disparity, which make it attract more and more attentions. Stereo-based 3D detection methods extract the implicit depth information of stereo images in different ways. \cite{no4} focuses on encoding object size prior, ground-plane prior, and depth information into an energy function to generate 3D proposals. \cite{stereorcnn} converts the 3D object detection problem to left and right 2D object detection and keypoint prediction, then uses geometric constraints to build the 3D detection box. \cite{pl, no37, no25} convert the estimated disparity map of the stereo image into pseudo LiDAR points, then use LiDAR-based methods to estimate the three-dimensional bounding boxes. These methods achieve the most advanced performance on the stereo image but cannot detect in real time because it is time-consuming to estimate the depth of entire image. Some traditional methods \cite{marapane1989region, wang2008region} use region-based stereo matching to estimate object's depth, but their performance is poor because they are not based on deep learning. \cite{ida3d} predicts the depth of the target through the method of instance depth perception which can detect the three-dimensional box end-to-end, but it neglects the local structure information. Our approach introduces a structure-aware-depth-estimation module that directly predicts the depth of the 3D bounding box’s center, and then rectifies the results by box estimation and dense alignment, which together benefit the accuracy of depth estimation and thus yield better 3D detection performance.

\section{SIDE}

\subsection{Overview}
Given the input stereo RGB images, our goal is to predict 3D bounding boxes and category labels for each object of interest. The attributes of predicted bounding box include the position of the object center (x, y, z), the three-dimensional size (w, h, l), and the steering angle $\theta$. Our method only needs the labels of 3D bounding boxes as supervision for each image.

The complete framework is shown in Fig.\ref{fig1}. For stereo detection, we associate the position of objects in the left and right images through the heatmap of objects’ center. The details will be introduced in Section \ref{sub:stereo}. For instance depth estimation, we only pay attention to the depth information of the center point of each object, so we construct local cost volume based on the RoIs of objects. Since the target depth information has strong sparsity and locality in the local cost volume, we make use of the structure and internal similarity of local cost volume to aggregate features. As a result, match reweight and structure-aware attention are introduced to make the information more concentrated and thus enhance the accuracy of depth estimation. The details will be introduced in Section \ref{sub:depth}. Finally, given the preliminary accurate estimated depth, we devise an efficient geometric post-processing scheme with the 3D-2D projection formula to further correct objects' positions. The details are in the Section \ref{sub:post}.

\subsection{Stereo Detection}
\label{sub:stereo}

\noindent\textbf{Stereo 2D Detection}\quad Similar to \cite{no40}, we use the heatmap of object’s center to represent each object. The heatmap is generated through the backbone and other branches are linked according to the position of each object in the heatmap. According to correspondence, the respective bounding boxes of each object in the stereo pictures can be detected at the same time. 

\begin{figure}[h]
    \centering
    \includegraphics[scale=1.0]{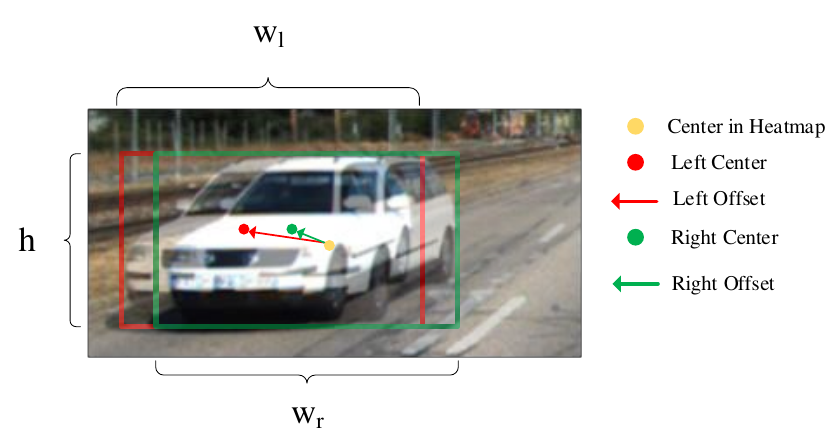}
    \caption{The relationship between left image and right image through center heatmap.}
    \label{fig_off}
\end{figure}

Usually, object's position is corrected in stereo pictures to ensure the vertical position of the same object is the same, so $ y1 $ and $ y2 $ in the bounding box of the stereo image are the same and we only need to predict a shared $ y $ when predicting the position. Therefore, the detection of the object's stereo 2D box is completed by the heatmap and two other branches, whose predictions are $ (o_{ul}, o_{ur}, o_v) $, $ (w_l, w_r, h) $. As shown in Fig.\ref{fig_off}, $ o_{ul} $ and $ o_{ur} $ represent the respective offset from the horizontal $ u $ coordinate of the object's center in the heatmap to stereo images, and $ o_v $ represents the offset of vertical $ v $ coordinate. $ w_r $ and $ w_l $ represent the respective widths of object's stereo 2D box, $ h $ represents the same height of the 2D box. 

For the generation of the image center heatmap, we use the same method in CenterNet~\cite{no40} and set $Y \in [0, 1]^{\frac{W}{R} \times \frac{H}{R} \times C}$ to label the image, where $W$ represents the width of image, $H$ represents the height of image, $R$ represents the multiple of downsampling, and $C$ represents the type of classification. In order to alleviate the imbalance problem of positive and negative samples, we use focal loss~\cite{lin2017focal} to train this network. Since the network down-sampled the image by $ R $, when the feature map is remapped to the original image, it will cause errors. Therefore, an additional offset needs to be predicted for each center and we predict the offset from the object's center in the heatmap to its center in the left and right image. In addition, we regress the length and width of the 2D stereo detection boxes after downsampling which are calculated in advance. All of these values are trained with L1 loss.

\noindent\textbf{Three-dimensional Attributes Detection}\quad In order to construct the 3D bounding box of the object, we need extra branches to predict the three-dimensional attributes of each object. Therefore, we also make predictions for the object's dimensions, steering angle, and keypoints. The dimensions include the length, width and height of 3D box, and keypoints include object's perspective keypoints~\cite{stereorcnn} and visible keypoints. Perspective keypoints used for constructing 3D-2D projection can help correct the steering angle and visible keypoints used for further dense alignment can help correct the location of 3D object in the post-processing. 

For the prediction of object's dimensions, because 3D dimensions of an object are three scalars, We directly regress to their absolute values in meters using a separate head. For the prediction of the steering angle $\theta$ of the object, we predict the allocentric angle $\alpha$ of each object instead of directly predicting the egocentric steering angle $\theta$. The two angles can be transformed by $\theta = \alpha + \arctan(\frac{x}{z})$. To avoid the discontinuity, the training targets are $ [\sin \alpha, \cos \alpha] $ instead of the raw angle value. both values are trained with L1 Loss.


\subsection{Structure-Aware Instance Depth Estimation}
\label{sub:depth}

\begin{figure*}
	\centering
	\includegraphics[scale=0.3]{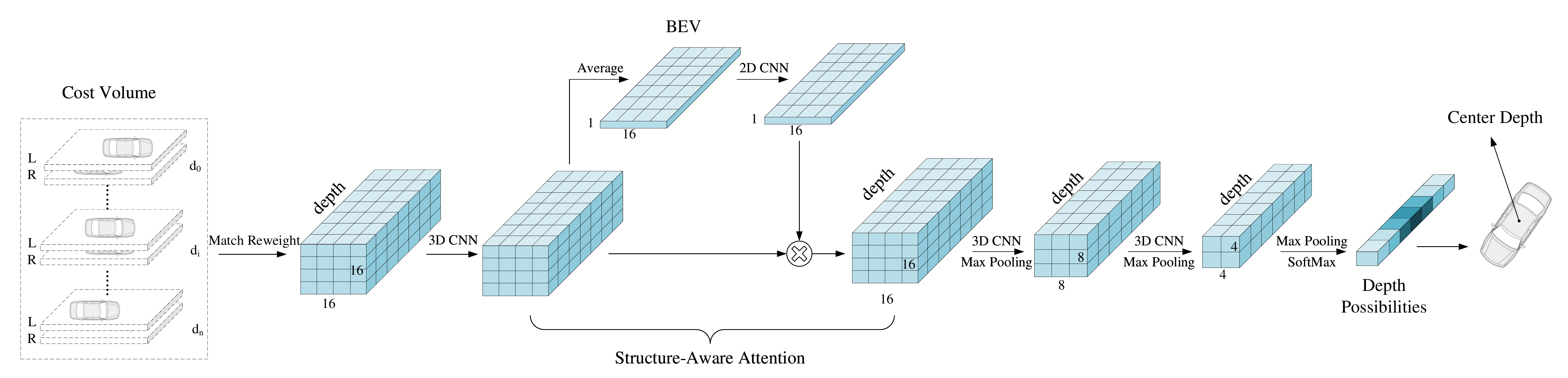}
	\caption{ Depth estimation module builds a 4D cost volume and performs 3DCNN to estimate the depth of a 3D bounding box center. We use two modules Match Reweight and Structure-Aware Attention to make information more focused and thus beneficial for estimating the depth.}
	\label{fig3}
\end{figure*}

Depth estimation is a very challenging problem in 3D detection. Next, we will elaborate the details of our devised structure-aware instance depth estimation module for tackling this problem in stereo-based 3D detection. Most previous work\cite{pl, no37, no25} construct cost volume of entire picture to calculate the disparity relationship corresponding to every pixel, then transform it to get the entire picture's depth information, which requires the use of depth map or disparity map for supervision during the training process. Compared with the previous, our depth estimation only pays attention to the information of the center depth of each detected object, and only needs to estimate the center depth of the object. Therefore, the cost volume needed to be constructed is smaller, and only the annotated object depth information in the 3D ground truth is required for supervision.

The detail of structure-aware instance depth estimation module is shown in Fig.\ref{fig3}. We use RoI Align\cite{2017Mask} to select the area of target object on the feature map of the stereo images, build 4D cost volume from the selected feature, and then feed it into a 2-stage 3D convolutional network. After reweighting the cost volume by calculating the similarity of the feature area, it will be fed into the first-stage 3D convolutional network, where we introduce a structure-aware attention mechanism to make the information more concentrated. In the second stage, we use Max Pooling to perform 2 times downsampling after passing 3D convolutional network twice respectively, then we use Max Pooling to perform 4 times downsampling and SoftMax to normalize the prediction. Relying on the network's normalization, the down-sampled features are finally merged into depth probability of the 3D box center. Therefore, the final result of the object's center depth can be calculated by $\hat{z} = \sum_{i = 1}^{N} z_i \times P(i)$, where $ N $ denotes the number of depth levels and $ P(i) $ is the normalized probability. We train our model with supervised learning using ground truth depth of 3D box center, where supervised regression loss is defined using the error between the ground truth depth $ z $ and the model’s predicted depth $ \hat{z} $

\begin{equation}
L_{depth} = \frac{1}{N} \sum^N_{k = 1} | z^{(k)}  - \hat{z}^{(k)} |. 
\label{eq6}
\end{equation}

\noindent\textbf{Construction of Cost Volume}\quad We use the depth as the dividing basis to ensure that the depth range is evenly divided, then convert the uniform depth range into the non-uniform disparity range, and construct the cost volume with the non-uniform disparity range. Disparity $d$ can be converted to depth $z$ by $ z = \frac{f_u \times b}{d} $, where $f_u$ represents horizontal focal length and $b$ represents the baseline of stereo camera. This equation shows that disparity and depth are in an inverse relationship, which means if the disparity is directly used as the dividing basis of cost volume, the far area will be insufficiently divided leading to inaccurate depth estimation of distant objects. Our construction of cost volume based on depth ensure we are also more likely to obtain accurate depth about objects at far location. 

In addition, since we estimate the depth for each specific object individually,  so we do not need to construct cost volume in such a large depth range. According to the width of the 2D box in stereo image, we can roughly calculate a more precise depth range for depth estimation by the camera intrinsic parameters.

\noindent\textbf{Match Reweight Strategy}\quad When calculating the predicted depth, we will perform a weighted average according to the normalized possibility of all depths in the cost volume range to get final result, rather than the depth value represented by the maximum probability of the cost volume, which ensures the continuity of the depth estimation range. However, when the difference in depth possibility is not obvious, the depth result will lack discriminativeness. Therefore, when constructing the cost volume, we introduced correlation scores to reweight the cost volume sequence. As shown in Eq.\ref{eq8}, the correlation score $ s $ at depth level $i$ is obtained by calculating the correlation between left and right feature maps in cost volume, where $F_l^{(i)}$ and $F_r^{(i)}$ are the pair of feature maps at corresponding depth level $i$.

\begin{equation}
s^{(i)} = \cos <F_l^{(i)}, F_r^{(i)}> = \frac{F_l^{(i)} \cdot F_r^{(i)}}{||F_l^{(i)}|| \times ||F_r^{(i)}||}. 
\label{eq8}
\end{equation} 

This equation uses cosine function to calculate the similarity of the corresponding regions of the feature map in the cost volume of each level. Then we use this similarity to reweight corresponding depth level in the cost volume. Our reweight method mainly comes from the idea that, in the construction of cost volume at different depth levels, the contents contained in the corresponding selected feature maps are different. When the similarity between the two feature maps is high, it means that there are more corresponding regions in the two feature maps at this depth level, which also means that this level is more likely to represent the center depth of the object.

\noindent\textbf{Structure-aware Attention Mechanism}\quad Although the match reweight strategy makes the cost volume's depth level more discriminant, there is also lots of spatial information noise in the 4D information space due to the sparsity of local cost volume. When we estimate the depth, we only estimate the depth of the center point of the object, which means the whole information space of cost volume is not well used. Inspired by some LiDAR-based methods\cite{no5, no18, no13}, which convert the intermediate feature space to the perspective of front view or bird's-eye-view to reduce the interference of unstructured spatial noise, we design a structure-aware attention module for the stereo image based 3D detection.


As shown in the structure-aware attention part of Fig.\ref{fig3}, after feeding the local cost volume into a 3D convolutional network, we averaged it on the Y-axis to get a depth feature space in 2D bird's-eye-view, then we use a 2D convolution and Sigmoid function $ \sigma $ to determine which parts in the feature space of bird's-eye-view are useful. Finally, the convolution result is multiplied by the original cost volume to achieve the effect of reducing  space noise. The attention process can be expressed by Eq.\ref{eq_att}

\begin{equation}
G_a = \sigma (Conv(Avg (G_h, dim = 2))) \otimes G_h + G_h
\label{eq_att}
\end{equation}

\subsection{Refinement of 3D Posture}
\label{sub:post}

After obtaining object's 2D box, 3D dimension, steering angle, and the depth of object's center, object's 3D position can be roughly calculated. Subsequently, we combine the objects' predicted perspective keypoints and visual keypoints to further correct the 3D position. In the post-processing, we use box estimation and dense alignment to get more accurate results.

\noindent\textbf{Geometric 3D Box Estimation}\quad According to \cite{stereorcnn, 2018Stereo}, given the left-right 2D boxes, perspective keypoint, and regressed dimensions, the 3D box center $ (x, y, z, \theta) $ can be solved by minimizing the projection error of 2D boxes and the keypoint. Since our network has a depth estimation module to get an accurate depth, there is no need to estimate the depth through the box estimator. Therefore, our box estimator is simpler than the box estimator adopted by Stereo R-CNN, we only need to use the 2D box information of the left picture, combined with the preliminary predicted depth $z$ and the keypoints to correct the data of $ x $, $ y $ and $\theta$. the projection formula of 3D-2D we formed is shown in Eq.\ref{eq10}. We extract five measurements from 2D boxes and perspective keypoints $ (u_l, v_t, u_r, v_b, u_p) $, which represent left, top, right, bottom edges of the left 2D box, and the $u$ coordinate of the perspective keypoint. $ w $, $ h $ and $ l $ represent the regression size, and $ x $, $ y $, $ z $ represent the coordinates of the center point of the 3D bounding box. 



\begin{equation}
\begin{cases}
u_l = (x - \frac{w}{2} \cos \theta - \frac{l}{2} \sin \theta) \ / \ (z + \frac{w}{2} \sin \theta - \frac{l}{2} \cos \theta) \\ 
v_t = (y - \frac{l}{2}) \ / \ (z + \frac{w}{2} \sin \theta - \frac{l}{2} \cos \theta) \\
u_r = (x + \frac{w}{2} \cos \theta + \frac{l}{2} \sin \theta) \ / \ (z - \frac{w}{2} \sin \theta + \frac{l}{2} \cos \theta) \\ 
v_b = (y + \frac{l}{2}) \ / \ (z - \frac{w}{2} \sin \theta + \frac{l}{2} \cos \theta) \\
u_p = (x + \frac{w}{2} \cos \theta - \frac{l}{2} \sin \theta) \ / \ (z - \frac{w}{2} \sin \theta - \frac{l}{2} \cos \theta) \\ 
\end{cases}
\label{eq10}
\end{equation}

Note that with regard to the perspective keypoints, only one of the bottom corners can be projected within the vertical border of the 2D box visually and the keypoints observed in different perspectives about the same car are different. Therefore, when predicting key points, we not only predict the distance of the keypoint relative to the border, but also predict the type of keypoint. In addition, when the object is truncated on the image, We use the viewpoint angle $\alpha$ to compensate the unobservable states.

\begin{figure}[h]
	\centering
	\includegraphics[scale=0.18]{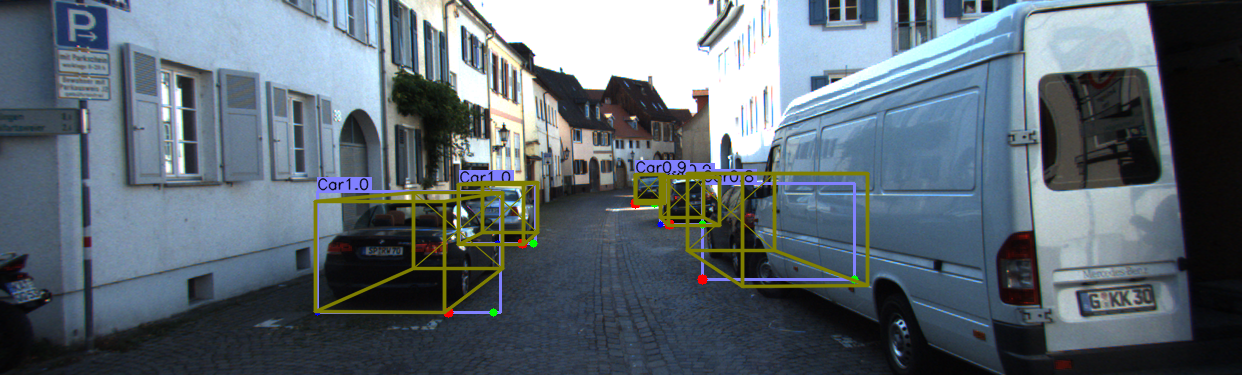}
	\caption{The red dot represents the left boundary of the visible range, and the green dot represents the right boundary of the visible range. It can be seen from the figure that the prediction of the visible range is inaccurate for cars with severe occlusion.}
	\label{fig6}
\end{figure}

\noindent\textbf{Dense 3D Box Alignment}\quad Within visual range from the predicted visual keypoints, we can sample the pixel of the object from the stereo images and calculate the corresponding error of each pixel at a given center depth, then we can obtain a more accurate depth by minimizing the sum of pixel's error. In practice, as shown in Fig.\ref{fig6}, we find that the prediction of the visual range about some highly occluded objects will be inaccurate, resulting in the center depth of the object after dense alignment is even more inaccurate, so we improve the original dense alignment module. For objects with a large occlusion, the visible range will be further reduced to ensure that the sampled points are from the object's 3D box, not from other objects that cover it. In addition, since the depth estimation can predict relatively accurate depth information, the depth range in the dense alignment can be reduced to speed up this module.

\section{Experiments}

\begin{table*}[]
\caption{Average precision of bird’s eye view $ AP_{BEV} $ and 3D boxes $ AP_{3D} $, evaluated on the KITTI validation set, where S denotes stereo image as input, M denotes monocular image as input and D denotes extra depth map as supervision.}
\label{table1}
\begin{spacing}{1.15}
\begin{tabular}{|c|c|c|ccc|ccc|}
\hline
\multirow{2}{*}{Method} & \multirow{2}{*}{Data} & \multirow{2}{*}{Time} & \multicolumn{3}{c|}{$ AP_{BEV} $ / $ AP_{3D} $(IoU=0.5)} & \multicolumn{3}{c|}{$ AP_{BEV} $ / $ AP_{3D} $(IoU=0.7)} \\ \cline{4-9} 
                        &                       &                       & Easy         & Mode        & Hard        & Easy         & Mode        & Hard        \\ \hline
M3D-RPN~\cite{brazil2019m3d}                 & M                     & 160ms                 & 55.37/38.96    & 42.49/39.57    & 35.29/33.01   & 25.97/20.27 & 21.18/17.06          & 17.09/15.21 \\
MonoPair~\cite{no6}                & M                     & 57ms                  & 61.06/55.38            & 47.63/42.39           & 41.92/37.99           & 24.12/16.28            & 18.17/12.30           & 15.76/10.42           \\
RTM3D~\cite{no17}                   & M                     & 55ms                  & 57.47/54.36    & 44.16/41.90    & 42.31/35.84   & 25.56/20.77 & 22.12/16.86          & 20.91/16.63\\ \hline
\demph{PL+FP}~\cite{pl}                   & \demph{S+D}                   & \demph{670ms}                 & \demph{89.80/89.50} &  \demph{77.60/75.50} & \demph{68.20/66.30} & \demph{72.80/59.40} & \demph{51.80/39.80} & \demph{44.00/33.50}          \\
\demph{PL+AVOD}~\cite{pl}                 & \demph{S+D}                   & \demph{510ms}                 & \demph{88.50/76.80} & \demph{76.40/65.10} & \demph{61.20/56.60} & \demph{61.90/60.70} & \demph{45.30/39.20} & \demph{39.00/37.00}          \\
\demph{PL++}~\cite{no37}                    & \demph{S+D}                   & \demph{500ms}                 & \demph{89.00/89.00} & \demph{77.50/77.80} & \demph{68.70/69.10} & \demph{74.90/63.20} &  \demph{56.80/46.80} & \demph{49.00/39.80}          \\
\demph{DSGN}~\cite{2020DSGN}                    & \demph{S+D}                   & -                     & -            & -           & -           & \demph{83.24/72.31} & \demph{63.91/54.27} & \demph{57.83/47.71}           \\ \hline
3DOP~\cite{no3}                    & S                     & -                     & 55.04/46.04    & 41.25/34.63    & 34.55/30.09   & 12.63/6.55  & 9.49/5.07            & 7.59/4.10\\
S-RCNN~\cite{stereorcnn}             & S                     & 417ms                 & 87.13/85.84    & 74.11/66.28    & 58.93/57.24   & 68.50/54.11 & 48.30/36.69          & 41.47/31.07 \\
IDA-3D~\cite{ida3d}                  & S                     & -                     & 88.05/87.08    & \textbf{76.69/74.57}    & 67.29/60.01   & 70.68/54.97 & 50.21/37.45          & 42.93/32.23\\ \hline
\algorithmname(R=4)               & S                     & 210ms                & 86.41/85.29    & 74.43/67.21    & 66.45/59.05   & 68.75/56.74 & 51.21/41.83 & 44.97/35.67           \\
\algorithmname(R=2)               & S                     & 260ms                 & \textbf{88.35/87.70}    & 76.01/69.13    & \textbf{67.46/60.05}   & \textbf{72.75/61.22} & \textbf{53.71/44.46} & \textbf{46.16/37.15 }          \\ \hline
\end{tabular}
\end{spacing}
\end{table*}


\noindent\textbf{Backbone}\quad Like the implementation in~\cite{no40}, we use dla-34~\cite{2018Deep} as the backbone of the network, and all hierarchical aggregation connections are replaced by DCN~\cite{dai2017deformable}. The number of channels output by dla34 is 64 and five of the detection branches are connected to the backbone network through two convolutions of sizes $ 3 \times 3 \times 64 $ and $ 1 \times 1 \times n $, where $ n $ is the characteristic channel of the relevant output branch. For the detection of keypoints, since it has more output, we deepen this branch to get more accurate results. As for the branch of the depth prediction, we first reduce the dimension of the stereo feature map through a convolution, then construct local cost volume to predict the center depth of each object through the instance depth estimation.

\noindent\textbf{Training}\quad We implement the models in PyTorch~\cite{2019PyTorch} and define the loss function of this multitask training by Eq.~\ref{eq11}. Each loss is weighted by their uncertainty following~\cite{kendall2018multi}. In addition, We double the training set by flipping the training set image horizontally and swap left and right images. We also adopts random clip and scale data augmentation for training and set the probability of random clip and scale to 0.35 to prevent excessive information from overflowing the image. We use the Adam~\cite{kingma2014adam} optimizer on 4 NVIDIA Tesla V100 GPU for 80 epochs of training. The initial learning rate is set to $ 2.5 \times 10^{- 4} $. The learning rate is reduced by 10 times at $25^{th}$, $40^{th}$, $60^{th}$ and $70^{th}$ epoch respectively. The backbone network is initialized by a classification model pretrained on ImageNet~\cite{2009ImageNet}. we train with 16 batch sizes of each GPU for about 9 hours.

\begin{equation}
	\begin{split}
	L = &w_{cls}L_{cls} + w_{off}L_{off} + w_{size}L_{size} + w_{dim}L_{dim} \\
	&+ w_{\theta}L_{\theta} + w_{kpts}L_{kpts} + w_{depth}L_{depth}. 
	\end{split}
	\label{eq11}
\end{equation}

\noindent\textbf{Evaluation}\quad We evaluate our method on the KITTI object detection benchmark~\cite{no1}. Following the same training and validation splits as~\cite{no4}, we devide 3712 images into traing set and 3769 images into validation set respectively. We report 3D average precision $ AP_{3D} $ and birds-eye-view average precision $ AP_{BEV} $ on car category with the IoU thresholds at 0.5 and 0.7. The category of car is divided into easy, moderate, and hard case according to the 2D box height, occlusion and truncation levels. 


\subsection{3D Detection Performance on KITTI}
\label{sub:per}

We conduct experiments both qualitatively and quantitatively. For comparison, we summarize the results mainly into two groups, monocular-based and stereo-based methods, then we set the downsampling factor R = 2 and R = 4 respectively to train the model and evaluate the performance of our 3D detection method by Average Precision for bird’s eye view $ AP_{BEV} $ and 3D box $ AP_{3D} $ as shown in Tab.\ref{table1}.

Compared with stereo-based methods without depth map supervision, we obtain the highest $ AP_{3D} $ and $ AP_{BEV} $. Specifically, we outperform 3DOP over 30\% for both $ AP_{BEV} $ and $ AP_{3D} $ across all kinds of cases. The performance of our model is also much better than Stereo R-CNN in the $ AP_{BEV} $ and $ AP_{3D} $ because Stereo R-CNN calculates the center depth of the object through a geometric method instead of directly predicting the center depth of the object, which causes inaccurate depth prediction. For IDA-3D, since this method will estimate the depth of object center, its performance is also better than Stereo R-CNN. But our method introduces the match reweight and attention mechanism to make the information more aggregated as well as uses geometric and pixel-level constraints to refine detection results in the post-processing, so the performance of our method are also comparable with IDA-3D, and it is better than IDA-3D in the easy and hard case with IoU=0.5 and all kinds of cases with IoU=0.7. 

We also list some stereo-based methods that require depth map as supervision. Although some of them perform better than our method, they all run much slower than our method. Compared with previous monocular-based methods, although the running time of these methods is lower than ours, our method outperforms previous monocular-based methods by a significant margin in all kinds of cases. These comparisons show that our method achieves a good balance between performance and efficiency. Furthermore, we report evaluation results on the KITTI testing set in Tab.\ref{tab_test}, compared with the testing set results of Stereo-RCNN, our method also shows superiority especially on the hard class.

\begin{table}[]
\caption{The performance of 3D detection on the KITTI testing set}
\label{tab_test}
\centering
\begin{tabular}{c|ccc}
\hline
\multirow{2}{*}{Method} & \multicolumn{3}{c}{\quad $AP_{3D}$(IoU=0.7) \quad}                \\ \cline{2-4} 
                        & Easy           & Mode           & Hard           \\ \hline
Stereo-RCNN             & 47.67          & 30.23          & 23.72          \\ \hline
\algorithmname                    & \textbf{47.69} & \textbf{30.82} & \textbf{25.68}\\
\hline
\end{tabular}
\end{table}

\subsection{Ablation Study}
\label{sub:as}

We conduct some ablation experiments to show the contribution of our proposed network modules.These experiments are performed on the car category in the KITTI dataset and we set the downsampling factor R = 4.

\noindent\textbf{Depth Estimation and Post-processing}\quad We first conduct experiments to verify the effect of instance depth estimation and geometric post-processing on the performance of 3D detection. The experimental results are shown in Tab.\ref{tab0}. When the instance depth estimation is not used, we directly use the center point disparity of the stereo 2D detection box to calculate the center depth, which leads to inaccurate 3D detection results. After adding instance depth estimation module, it shows that the performance of 3D detection especially in $AP_{BEV}$ has been greatly improved, and the final performance can be further improved through geometric post-processing.

\begin{table}[]
\caption{Contribution of instance depth estimation and geometric post-processing}
\label{tab0}
\footnotesize
\begin{spacing}{1.2}
\begin{tabular}{c|ccc}
\multirow{2}{*}{config}                                                          & \multicolumn{3}{c}{$ AP_{BEV} $ / $ AP_{3D} $(IoU=0.5)}                            \\ \cline{2-4} 
                                                                                 & Easy                 & Mode                 & Hard                 \\ \hline
w/o Estimation                                                             & 54.24/32.02          & 41.84/26.77          & 39.95/26.82          \\
\hline
w/ Estimation                                                              & 66.73/32.15          & 58.20/31.68          & 51.90/28.96          \\
\hline
\begin{tabular}[c]{@{}c@{}}w/ Estimation\\ w/ Post-processing\end{tabular} & \textbf{86.41/85.29} & \textbf{74.43/67.21} & \textbf{66.45/69.05}
\end{tabular}
\end{spacing}
\end{table}

\noindent\textbf{Match Reweight and Attention Module}\quad During depth estimation, correlation score reweights cost volume according to the similarity of different depth levels and structure-aware attention reduces the noise in the original space by the information from bird's eye view of the object. The purpose of these two modules is to improve the possibility of correct depth and make the result of depth estimation more discriminant. As is shown in Tab.~\ref{table2}, the performance of our method can be improved by combining these two modules.

\begin{table}[h]
	\caption{Improvements of match-reweight and structure-aware attention, where Re. represents match reweight and Att. represents structure-aware attention.}
    \begin{tabular}{|cc|ccc|}
		\hline
		\multirow{2}{*}{Att.} & \multirow{2}{*}{Re.} & \multicolumn{3}{c|}{$ AP_{BEV} $ / $ AP_{3D} $ (IoU=0.5) }                                                                         \\ \cline{3-5} 
		&                              & Easy                            & Mode                            & Hard                             \\ \hline
		&                              & 84.39/83.45	&	72.93/65.71 &	65.08/57.78                     \\
		$\checkmark$                     &                              & 
		85.84/84.84     &       73.88/66.91     &       66.02/58.80                      \\
		\multicolumn{1}{|l}{} & $\checkmark$                             & \multicolumn{1}{l}{85.46/84.63} & \multicolumn{1}{l}{73.95/66.81} & \multicolumn{1}{l|}{66.28/58.86} \\
		$\checkmark$                      & $\checkmark$        
        &       \textbf{86.41/85.29}     &	\textbf{74.43/67.21}     &	\textbf{66.46/59.04}                     \\ \hline
	\end{tabular}
	\label{table2}
\end{table}

\noindent\textbf{Box Estimator and Dense Alignment}\quad Although the coarse 3D box has a precise projection on the image because of the depth estimation, it is not accurate enough for 3D localization. Therefore, we need to correct the steering angle of 3D box through box estimation and refine the depth of 3D box through dense alignment. The results in Tab.\ref{table3} show that the performance of our method is improved through these two steps. Note that in the process of dense alignment, we need to use box estimation again after fixing the depth to recover the 3D box. In addition, we find that the improvement of 3D detection is limited when only dense alignment is used, because it requires 3D boxes to fit closely with objects, while the box estimator can make these 3D boxes fit more closely with corresponding objects by correcting the steering angle.

\begin{table}[h]
	\caption{Improvements of dense alignment and 3D box estimation, where Est. represents 3D box estimation and Ali. represents dense alignment. .}
    \begin{tabular}{|cc|ccc|}
		\hline
		\multirow{2}{*}{Est.} & \multirow{2}{*}{Ali.} & \multicolumn{3}{c|}{$ AP_{BEV} $ / $ AP_{3D} $ (IoU=0.5) }                                                                         \\ \cline{3-5} 
		&                              & Easy                            & Mode                            & Hard                             \\ \hline
		&                              & 66.73/32.15 & 58.20/31.68 & 51.90/28.06                     \\
		$\checkmark$                     &                              & 
        81.35/74.61 & 64.29/61.93 & 56.81/54.41
                    \\
		\multicolumn{1}{|l}{} & $\checkmark$                             & \multicolumn{1}{l}{74.88/69.46} & \multicolumn{1}{l}{62.00/58.30} & \multicolumn{1}{l|}{54.83/51.85} \\
		$\checkmark$                      & $\checkmark$                             & \textbf{86.41/85.29} &	\textbf{74.43/67.21} &	\textbf{66.45/59.05}                      \\ \hline
	\end{tabular}
	\label{table3}
\end{table}

\noindent\textbf{The Benefit of Trainig Strategy}\quad We use two strategies to enhance model performance during the training stage, image flip augmentation and weight uncertainty. We conduct different combinations of experiments on these two strategies and the results are shown in Tab.~\ref{table4}. Weight uncertainty can balance the multi-task loss and avoid manual adjustment of weights. The data flip augmentation doubles the number of samples to achieve better accuracy. Both of the strategies make our method achieve better performance.

\begin{table}[H]
	\caption{Improvements of using data flip augmentations and uncertainty weight.}
    \small
    \begin{tabular}{|cc|ccc|}
		\hline
		\multirow{2}{*}{Flip} & \multirow{2}{*}{{\footnotesize Uncertainty}} & \multicolumn{3}{c|}{$ AP_{BEV} $ / $ AP_{3D} $ (IoU=0.5) }                                                                         \\ \cline{3-5} 
		&                              & Easy                            & Mode                            & Hard                             \\ \hline
		&                              & 77.16/75.52	& 62.48/60.32 &	55.11/53.29                     \\
		$\checkmark$                     &                              & 
		79.23/77.42	& 64.53/62.40 &	57.03/55.31                      \\
		\multicolumn{1}{|l}{} & $\checkmark$                             & \multicolumn{1}{l}{80.13/77.19} & \multicolumn{1}{l}{63.79/61.90} & \multicolumn{1}{l|}{56.42/54.38} \\
		$\checkmark$                      & $\checkmark$                             & \textbf{86.41/85.29}    & \textbf{74.43/67.21}    & \textbf{66.45/59.05}                     \\ \hline
	\end{tabular}
	\label{table4}
\end{table}

\noindent\textbf{Qualitative Results}\quad We show the qualitative results of some scenarios from KITTI dataset in the supplementary materials.


\section{Conclusion}


In this paper, we propose a center-based stereo 3D detection method which has better performance especially when detecting objects in hard condition (such as farther from the camera and more severe occlusion). Since depth information is essential for 3D detection, we estimate object’s center depth via constructing local cost volume from its RoI. We also introduce match-reweight and structure-aware attention to aggregate information and reduce space noise caused by information sparsity. By overcoming this sparsity, the center of object can be predicted more accurately. In addition, we use the anchor-free model to speed up objects’ 2D detection and further improve the accuracy of results in the post-processing through object’s geometric and pixel-wise constraint. By predicting more accurate location of keypoints, the corrected result can be more accurate.

{\small
\bibliographystyle{ieee_fullname}
\bibliography{egbib}
}

\end{document}